\def\year{2022}\relax
\documentclass[letterpaper]{article} 
\usepackage{aaai22}  
\usepackage{times}  
\usepackage{helvet}  
\usepackage{courier}  
\usepackage[hyphens]{url}  
\usepackage{graphicx} 
\usepackage{natbib}  
\usepackage{caption} 
\DeclareCaptionStyle{ruled}{labelfont=normalfont,labelsep=colon,strut=off} 
\usepackage{algorithm}
\usepackage{algorithmic}

\usepackage{newfloat}
\usepackage{listings}
\floatstyle{ruled}
\newfloat{listing}{tb}{lst}{}
\floatname{listing}{Listing}
\title{Pixel VQ-VAEs for Improved Pixel Art Representation}
\author {
    Akash Saravanan,
    Matthew Guzdial\\
}
\affiliations {
    University of Alberta \\
    akash.saravanan@ualberta.ca, guzdial@ualberta.ca
}

\begin{document}

\maketitle

\begin{abstract}
Machine learning has had a great deal of success in image processing. However, the focus of this work has largely been on realistic images, ignoring more niche art styles such as pixel art. Additionally, many traditional machine learning models that focus on groups of pixels do not work well with pixel art, where individual pixels are important. We propose the Pixel VQ-VAE, a specialized VQ-VAE model that learns representations of pixel art. We show that it outperforms other models in both the quality of embeddings as well as performance on downstream tasks.
\end{abstract}

\section{Introduction} 
Deep neural networks have been used for a variety of image-related tasks including image generation \cite{goodfellow-gan-2020}, transformation \cite{gonzalez-etal}, and translation \cite{image-to-image-2019}. However, the majority of this work focuses on photo-realism while avoiding other, more unrealistic art styles despite their usage in popular media. Pixel art is one such art style, characterized by a restricted color palette and discrete visible blocks of pixels (e.g. far-left of Table \ref{table:reconstructed_outputs}). Originally created for 8- \& 16-bit games, pixel art has remained popular, appearing in games like Minecraft, Pokémon, and Stardew Valley, as well as animations and webcomics \cite{pixel-art-animation,digital-storytelling}.
Improving our ability to work with pixel art in ML models could thus impact several domains. More specifically, this will lead to an improvement in any task that involves pixel art including generation and transformation of images in and outside of games.

Prior work on pixel art has focused on specific tasks, primarily that of image generation \cite{horsley-sprites-dcgan-2017,karth-2021-vqvae}. A shared representation for pixel art offers value as a common starting point for different tasks by saving human effort. In machine learning such representations, known as embeddings, are information-rich, multi-dimensional vectors learned by models for use in downstream tasks.
Understandably, the vast majority of prior work on learning embeddings focuses on photorealism, leaving other art styles, including pixel art, largely unexplored. 

In terms of learning embeddings, Variational Auto Encoders (VAEs) \cite{vae} dominate the field due to their excellent representational capabilities \cite{word2vec,gonzalez-etal}. However, the generated images can appear blurry and lack detail \cite{vae-blurry}, although recent work addresses this \cite{vq-vae,dai2019diagnosing}. Pixel art, characterized by discrete blocks of pixels, loses this defining property when the image is blurred. However, for pixel art, the problem may not be the VAEs themselves. Rather, Convolutional Neural Networks (CNNs), the backbone of many image processing models, process clumps of neighboring pixels together. This makes it difficult for such models to precisely target individual pixels in pixel art. One further complication is that pixel art is hand-authored and does not use natural images, thus severely limiting the available data.

We identify two drawbacks with current approaches for pixel art. First, prior work has approached individual tasks separately when a common representation could have been used. Second, CNNs do not natively work well with pixel art. To address the former, we propose a novel system to learn high quality representations of pixel art for use in multiple downstream tasks. For the latter, we introduce two new enhancements to improve CNN performance on pixel art.

Our contributions can be summarized as follows: 
\begin{itemize}
\item We propose the usage of VQ-VAEs \cite{vq-vae} to represent pixel art as a set of discrete embeddings, each mapped to groups of individual pixels. This focus on pixels synergizes well with pixel art where each individual pixel matters.
\item We introduce the Pixel VQ-VAE, which uses two key enhancements, the PixelSight block and the Adapter layer, to improve performance on pixel art.
\item We evaluate our Pixel VQ-VAE against several baselines on the quality of embeddings while also studying its performance on multiple downstream tasks.
\item We demonstrate the superiority of Pixel VQ-VAE on pixel art tasks, especially for a high variance dataset.
\end{itemize}

\section{Background}
In this section we cover prior work on image embedding approaches, their usage in pixel art as well as an overview of VQ-VAEs, an important component of our approach.

\subsection{Representation in Machine Learning}
Variational Auto Encoders (VAEs) \cite{vae} are commonly used to learn representations of content. Known as embeddings, these representations have achieved success across many domains \cite{word2vec}. However, in the case of images, VAEs tend to produce blurry and unclear outputs \cite{vae-blurry}. For pixel art in particular, this is a major problem as having discrete, visible pixels is an essential part of the art style. While several works have demonstrated success in enhancing the quality of reconstructions, they have not been adapted to unrealistic art styles such as pixel art \cite{dai2019diagnosing,larsen2016autoencoding}. 

There are several approaches to image generation that could be employed for pixel art. The most popular are Generative Adversarial Networks (GANs) \cite{goodfellow-gan-2020} and their variants that focus on improving the quality of generated images \cite{radford-2016-dcgan,image-to-image-2019}. These approaches generally require large corpora of data to achieve good results. However, most art styles (including pixel art) that don't use natural images have far less data available as they are usually hand-authored. This is further complicated in approaches like diffusion models \cite{glide} which require labeled data. Additionally, while diffusion models have been used to generate pixel art, the low resolution and varied structure of pixel art would make it difficult to generate sprites for specific games. Another class of image generation models—PixelCNNs—are auto-regressive in nature. That is, they generate images one pixel at a time, which makes their use for pixel art appealing \cite{pixel-rnn,conditional-pixel-cnn}. However, the learning problem grows more complex due to modeling relationships both between individual pixels and their color channels.
 
\subsection{Representation in Pixel Art}

Pixel art appears commonly in video games, thus work on representing video game content often requires representing pixel art. The Video Game Level Corpus \cite{vglc} represents game levels as a set of discrete symbols. \citet{mrunal-tile-embedding-2021} utilized a VAE to learn embeddings of game level components. \citet{karth-2021-vqvae} used a VQ-VAE to learn encodings which were then used for map generation through a secondary process. These approaches do not focus directly on pixel art representation.
Closer to our work, \citet{gonzalez-etal} used a VAE to learn representations of Pokémon art along with associated gameplay-relevant properties that could then be modified to generate new Pokémon. However, this VAE suffered from the aforementioned blurriness issue. 

There are several approaches for tasks we consider downstream to learned embeddings. Both \citet{serpa-spritesheets-2019}, and Pokémon2Pokémon \cite{pokemon2pokemon} used GANs for coloring images. The former converted 2D line-art of sprites into colored versions while the latter altered color palettes. Palette modification is an application of our embeddings that we demonstrate later in this paper. \citet{horsley-sprites-dcgan-2017} used a DCGAN to generate new characters, another task we demonstrate. Additionally, while there are several open-source, unpublished works that work on Pokémon generation \citep{pokemon-aegan,pokemon-gan}, they primarily focus on the generation aspect and not the representations themselves.

\subsection{VQ-VAE}
Similar to a VAE, the Vector Quantized VAE (VQ-VAE) consists of an encoder, decoder and a latent space. However, unlike the VAE, it learns a discrete latent space. A key component to learning this is the presence of a codebook which maps discrete encodings (integers) to specific embeddings (vectors) in the latent space. The number of unique encodings learned is a hyperparameter while the embeddings are learned while training. The encoder takes in an image and outputs a grid of high-dimensional vectors. For each vectors, the closest codebook embedding is identified via a nearest neighbor search and the corresponding encoding is returned. The decoder converts the resultant grid of encodings back into an image. We refrain from a more detailed explanation due to space constraints but encourage readers to refer to the original paper \cite{vq-vae}. Like traditional embedding models \cite{word2vec}, a VQ-VAE is generally used to generate embeddings while a different model, such as a PixelCNN, uses this representation as the input to downstream tasks. We note that due to the discrete nature of the latent space, traditional methods of analysis such as t-SNE are not very useful.

We use a VQ-VAE as the basis of our work due to a special property it possesses - a direct correspondence between individual encodings and patches of pixels in the final image. This synergizes well with pixel art itself being composed of discrete blocks of pixels. To the best of our knowledge, we are the first to leverage this synergy for pixel art. 

\section{Pixel VQ-VAE}

\begin{figure*}[t]
    \centering
    \includegraphics[width=6.3in]{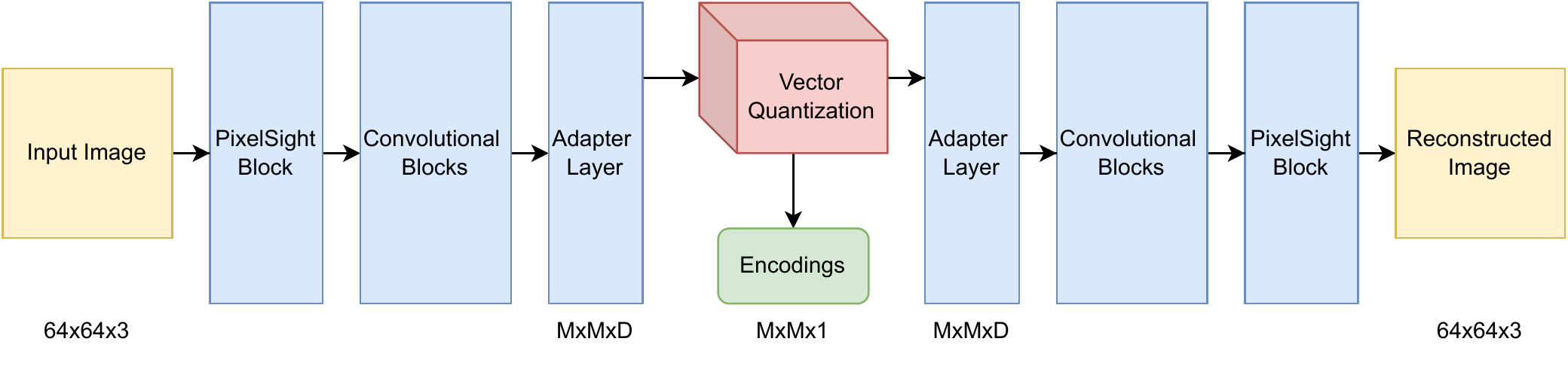}
    \caption{Pixel VQ-VAE Architecture}
    \label{fig:model_arch}
\end{figure*}

In this section we introduce our Pixel VQ-VAE, which makes use of two enhancements over a traditional VQ-VAE to better represent pixel art. Before we proceed with a detailed discussion of our enhancements, we overview some relevant properties and hyperparameters. Since the VQ-VAE performs a nearest neighbor search to discretize embeddings, it requires that the number of encoder output channels be equal to the dimensionality of the learned embeddings $D$. $K$ is the number of unique discrete encodings (integers) associated with those embeddings (vectors). Furthermore, we define an additional hyperparameter $M$, the encoding-pixel correspondence. This parameter is implicitly set via the formula $M = I / 2^L$ where $I$ is the size of the input and $L$ is the number of scaling convolutional blocks in the model. The encoder outputs embeddings of shape $I^2/M \times D$ which are discretized into $I^2/M$ encodings that are then converted back into a $I \times I \times 3$ image by the decoder. As an example, given a $64\times64\times3$ image, if we set $D=32$, $K=128$, and $L=2$, then $M=16$. This gives us $64 * 64 / 16 = 256$ embeddings of dimensionality $32$ to represent the entire image where each embedding corresponds to one of K ($128$) discrete encodings. We note that larger values of $L$ result in lower values of $M$. That is, the number of layers in the model and the number of encodings are inversely correlated.

We introduce two new enhancements - the ``PixelSight" block and the ``Adapter" layer. The PixelSight block uses 1x1 convolutions of stride 1, traditionally used for reducing model complexity \citep{1x1-paper}, as a means of giving the model an initially granular view of the input so that every pixel is considered individually. We use this alongside a batch normalization layer and a ReLU activation. We note that our PixelSight block is not restricted to only the VQ-VAE but can be plugged in to improve the pixel art performance of any convolutional model. Meanwhile, the Adapter layer solves a problem with the general VQ-VAE architecture. Specifically, the standard practice when using convolutional layers is to halve the input size and double the number of filters at each convolution. However, since the encoder is restricted to an output dimensionality of $D$, the final convolutional layer must have $D$ filters. We found that this parameter $D$ had significant impact on the model. Low values resulted in too few filters for the model to learn effectively while higher values led to giant models that had trouble converging. The Adapter convolutional layer uses a 1x1 with stride 1 like in traditional literature \citep{1x1-paper} to reduce the number of filters down to $D$ while retaining the remaining shape of the vector. This allows for the standard approach to be used for the preceding blocks. 

Figure \ref{fig:model_arch} illustrates our general model architecture. The encoder consists of the PixelSight block, $L$ convolutional blocks, and the Adapter layer. Each of the convolutional blocks consist of 2x2 convolutional layers with stride 2 like in \citet{horsley-sprites-dcgan-2017,gonzalez-etal}, a batch normalization layer and ReLU activation as per standard practice. The Adapter layer uses a linear activation function to leave the embedding space unrestricted. We calculate the number of filters for the remaining layers such that the final one has $F$ filters, with each preceding one having half as many. The decoder is a mirrored version of the encoder that starts with the Adapter layer to increase the number of filters back to $F$, followed by $L$ transposed convolutional blocks, a PixelSight block with a transposed convolutional layer and a sigmoid activation. 

\section{Experiments}
In this section we describe the implementational details of our Pixel VQ-VAE and baseline models. To understand if our work better models pixel art, we compare it against these baselines on the quality of the learned embeddings. We also perform an ablation study to validate our enhancements. Finally, we verify the usefulness of the learned embeddings on two common downstream tasks. We make all code publicly available for reproducibility.\footnote{\url{https://github.com/akashsara/fusion-dance}}

\subsection{Data}
All our experiments are performed on a dataset of sprites compiled from the Pokémon video game series  \cite{veekun-2017}. This dataset consists of monsters in a wide variety of shape, size and color, with no consistent features between them. We chose to use Pokémon not only due to its usage in prior work \cite{liapis2018recomposing}, but also due to the high variance present in the dataset. This variance is important since many models are capable of generating consistent structures in low variance environments but struggle with high variance. On another pixel art dataset \citep{sprites-dataset-2018} which comprises of highly consistent images, we found that our method still resulted in improvements, though to a lesser extent. We refrain from including these results in this work due to space constraints.

\subsection{Preprocessing}
We augment our data like \citet{gonzalez-etal} by using 4 different backgrounds - black \& white to handle some edge cases with some Pokémon, and for the training data alone, two randomly generated noisy backgrounds as regularization to help the model learn to ignore the background. We also perform horizontal flips and 4 random rotations (up to 30 degrees) in either direction. We acknowledge that these rotations may cause harmful effects in terms of pixel art style in exchange for a much larger dataset. However, this trade-off showed a significant improvement in all results for all models and we thus retain it. Due to different games having images of different sizes, we resized them to 64x64 using bicubic interpolation. While splitting the data, we ensure that no sprites of the same entity are duplicated in different splits. Our final dataset has 168,334 training, 3,046 validation and 6,999 test images. 

\subsection{Training}
We train 3 models of decreasing encoding-pixel correspondence by setting $M=16,4,1$ ($L=2,1,0$). As this results in more detail due to the larger number of encodings ($I^2/M$), we term the three models as Pixel VQ-VAE LowRes, MedRes and HiRes respectively with a note that the LowRes model has the most layers ($L=2$) and thus the most parameters. We empirically determined that the hyperparameters $K=256$, $F=512$, $D=32$ (MedRes and HiRes) and $D=64$ (LowRes) offered the best performance while still retaining a low value for $K$. All models use an Adam optimizer with learning rate 0.0001 and batch size 64. We use a Mean Squared Error (MSE) reconstruction loss term and train the models to convergence (25 epochs).

\subsection{Baselines}
Our first baseline is a standard VAE consisting of a mirrored encoder-decoder architecture each with 4 convolutional blocks. We train it to convergence (25 epochs) using the Adam optimizer with a batch size of 64 and a default learning rate (0.0001) on the standard training objective (Mean Squared Error + KL-Divergence). Additionally, we train and use LowRes and MedRes versions of a standard VQ-VAE which have none of our enhancements. We note that a HiRes version of the standard VQ-VAE is not feasible since it requires the number of scaling convolutional blocks to be $L=0$. The HiRes version is only possible for the Pixel VQ-VAE due to the enhancements we use.

\subsection{Embedding Quality}

\begin{table*}[t]
\centering
\begin{tabular}{ccccccc}
\hline
\textbf{Original} & \textbf{VAE} & \textbf{VQ-VAE} & \textbf{Pixel VQ-VAE} & \textbf{VQ-VAE} & \textbf{Pixel VQ-VAE} & \textbf{Pixel VQ-VAE}\\
\textbf{} & \textbf{} & \textbf{LowRes} & \textbf{LowRes} & \textbf{MedRes} & \textbf{MedRes} & \textbf{HiRes}\\
\hline\\
\includegraphics[]{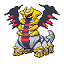} & \includegraphics[]{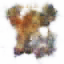} & \includegraphics[]{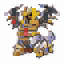} & \includegraphics[]{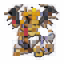} & \includegraphics[]{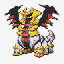} & \includegraphics[]{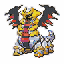} & \includegraphics[]{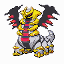} \\
\includegraphics[]{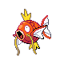} & \includegraphics[]{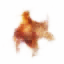} & \includegraphics[]{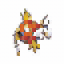} & \includegraphics[]{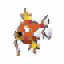} & \includegraphics[]{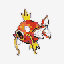} & \includegraphics[]{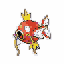} & \includegraphics[]{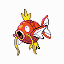} \\
\hline
\end{tabular}
\caption{Test set reconstructions of the Pixel VQ-VAE and the baselines. The VQ-VAE HiRes is not included as it cannot exist without the enhancements introduced in the Pixel VQ-VAE.}
\label{table:reconstructed_outputs}
\end{table*} 

\begin{table}[t]
\centering
\begin{tabular}{l|r|r}
\hline
\textbf{Model} & \textbf{MSE} & \textbf{SSIM}\\ 
\hline
VAE & 0.01815 & 0.64272 \\ 
\hline
VQ-VAE LowRes & 0.01076 & 0.60198 \\ 
Pixel VQ-VAE LowRes & 0.01040 & 0.78669 \\ 
\hline
VQ-VAE MedRes & 0.00584 & 0.63973 \\ 
Pixel VQ-VAE MedRes & 0.00421 & \textbf{0.91210} \\ 
\hline
Pixel VQ-VAE HiRes & \textbf{0.00070} & 0.82967 \\
\hline
\end{tabular}
\caption{Test set reconstruction metrics. MSE: Lower is better. SSIM: Higher is better. The VQ-VAE HiRes is not included as it cannot exist without the enhancements introduced in the Pixel VQ-VAE.}
\label{table:reconstruction_metrics}
\end{table} 

We evaluate all models in terms of MSE and Structural Similarity Index Metric (SSIM) of their reconstructions against the ground truth images of the test set. MSE is a per-pixel loss function which compares every pixel between the ground truth image and the reconstructed image while SSIM compares the structural similarity between two images based on statistical measures. We use these metrics as they complement each other well. MSE focuses on individual pixels while ignoring overall image structure while SSIM focuses on the structure of the image over the specific colors used. 

Table \ref{table:reconstructed_outputs} gives a visual comparison of the baseline models and our Pixel VQ-VAE. Visually, all VQ-VAEs surpass the VAE which is blurry and lacks detail. While both the VQ-VAEs and our Pixel VQ-VAEs do well, our Pixel VQ-VAE offers better detailing in the precision of the color palette. This is evident in Table \ref{table:reconstruction_metrics} which compares the performance of the different models. In all cases our Pixel VQ-VAE beats the baseline models, especially in terms of SSIM. Comparing our three Pixel VQ-VAEs, we see that the MSE is better for models with lower values of $M$. However, the MedRes model has a higher SSIM score than the HiRes version. This is because the HiRes model has a 1:1 encoding-pixel correspondence. This means that the model focuses heavily on individual pixels, and not overall structure. We note that despite this, both models far outperform all baselines. 

Since embeddings represent information about an entity, a good embedding generally exhibits characteristics of this information in the latent space. This information can be helpful in downstream tasks \cite{mrunal-tile-embedding-2021}. However, for a VQ-VAE, the latent space cannot be analyzed in the same manner. This is due to the fact that the learned embeddings correspond to individual patches of an image and not the image as a whole. But this does not mean that our Pixel VQ-VAE's embeddings are without use. Further in this paper we demonstrate the usefulness of these embeddings in two different downstream tasks.

\subsection{Ablation Study}

\begin{table}[t]
\centering
\begin{tabular}{l|l|r|r}
\hline
\textbf{Model} & \textbf{Size} & \textbf{MSE} & \textbf{SSIM}\\
\hline
Base VQ-VAE & LowRes & 0.01076 & 0.60198 \\ 
Adapter VQ-VAE & LowRes & 0.01049 & 0.77879 \\ 
PixelSight VQ-VAE & LowRes & 0.01079 & \textbf{0.79586} \\ 
Pixel VQ-VAE & LowRes & \textbf{0.01040} & 0.78669 \\ 
\hline
Base VQ-VAE & MedRes & 0.00584 & 0.63973 \\ 
Adapter VQ-VAE & MedRes & 0.00489 & 0.66650 \\ 
PixelSight VQ-VAE & MedRes & 0.00504 & 0.78298 \\ 
Pixel VQ-VAE & MedRes &\textbf{ 0.00421} & \textbf{0.91210} \\ 
\hline
PixelSight VQ-VAE & HiRes & 0.00133 & 0.73070 \\ 
Pixel VQ-VAE & HiRes & \textbf{0.00070} & \textbf{0.82967} \\ 
\hline
\end{tabular}
\caption{Ablation study comparing test set reconstruction metrics. MSE: Lower is better. SSIM: Higher is better. For HiRes models, the base VQ-VAE and the Adapter VQ-VAE cannot exist without further enhancements to the model.}
\label{table:ablation-pokemon}
\end{table} 

We next demonstrate an ablation study to verify the effectiveness of the enhancements used in our model. We compare our model to the baseline VQ-VAE as well as versions that use only one of our two enhancements. We denote these as ``Base" referring to the baseline, ``Adapter" referring to the use of the Adapter layer alone and ``PixelSight" referring to the use of the PixelSight block. We run this comparison for all versions of the Pixel VQ-VAE (LowRes, MedRes, HiRes). As described earlier, a HiRes version of the baseline VQ-VAE cannot exist without our enhancements. Similarly, the adapter requires at least one other layer in the model, thus a HiRes Adapter VQ-VAE does not exist either. 

Table \ref{table:ablation-pokemon} compares model performance on MSE and SSIM. Here, the enhanced VQ-VAEs always do better than the baseline VQ-VAE. Further, in nearly every case the Pixel VQ-VAE is the clear winner. The sole exception is the LowRes model where all three enhanced models exhibit very similar values. This can be attributed to the LowRes models having a large encoding-pixel correspondence ($M$=16). This leads to each individual encoding corresponding to a larger portion of the final image, thus leading to less control over the finer aspects. In the more controlled MedRes and HiRes cases the Pixel VQ-VAE significantly outperforms the other models. 

\subsection{Image Generation}

\begin{table*}[t]
\centering
\begin{tabular}{ccccccc}
\hline
\textbf{VAE} & \textbf{DCGAN} & \textbf{VQ-VAE} & \textbf{Pixel VQ-VAE} & \textbf{VQ-VAE} & \textbf{Pixel VQ-VAE} & \textbf{Pixel VQ-VAE}\\
\textbf{} & \textbf{} & \textbf{LowRes} & \textbf{LowRes} & \textbf{MedRes} & \textbf{MedRes} & \textbf{HiRes}\\
\hline\\
\includegraphics[]{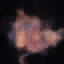} & \includegraphics[]{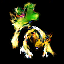} & \includegraphics[]{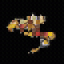} & \includegraphics[]{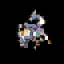} & \includegraphics[]{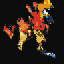} & \includegraphics[]{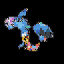} & \includegraphics[]{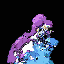}\\

\includegraphics[]{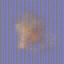} & \includegraphics[]{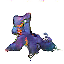} & \includegraphics[]{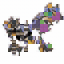} & \includegraphics[]{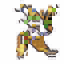} & \includegraphics[]{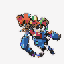} & \includegraphics[]{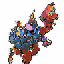} & \includegraphics[]{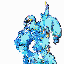}\\

\hline
\end{tabular}
\caption{Hand-picked generated samples for each model. All VQ-VAE results are generated from a PixelCNN model trained on the VQ-VAE embeddings.}
\label{table:generated_images}
\end{table*}

\begin{table}[t]
\centering
\begin{tabular}{l|r|r}
\hline
\textbf{Model} & \textbf{Parameters} & \textbf{FID Score}\\
\hline
VAE & 7,680,774 & 322.746\\ 
GAN & 15,424,000 & 101.759\\
\hline
VQ-VAE LowRes & 20,998 & 167.464\\
Pixel VQ-VAE LowRes & 1,413,065 & 154.735\\
\hline
VQ-VAE MedRes & 17,222 & 118.253\\
Pixel VQ-VAE MedRes & 1,102,121 & 124.302\\
\hline
Pixel VQ-VAE HiRes & 54,313 & \textbf{98.575}\\
\hline
\end{tabular}
\caption{FID of generated images. Lower is better. VQ-VAEs use PixelCNNs (13,444,864 parameters) for generation.}
\label{table:fid_metrics}
\end{table} 

Image generation is the most common downstream task for image embeddings. For our baselines, we use two common approaches to image generation, namely a VAE and a DCGAN \cite{horsley-sprites-dcgan-2017}. We use the same VAE from above and a standard DCGAN with increased parameters in the generator to have a similar number of parameters to our final model. The DCGAN is similar in architecture to unpublished works that have explored Pokemon generation \citep{pokemon-gan}. Both models were trained on the same dataset as our VQ-VAE. As discussed earlier, VQ-VAEs are like traditional word embedding models in that the model that learns the embeddings (VQ-VAE) is distinct from the model that uses the embeddings for downstream tasks. In our case we use the PixelCNN \cite{pixel-rnn}, a common downstream model used with VQ-VAEs \cite{vq-vae}. Specifically, we use a Gated Conditional PixelCNN \cite{conditional-pixel-cnn} which has several optimizations for better generation, including conditional image generation. The final model was trained on the embeddings generated by the Pixel VQ-VAE and conditioned on both the Pokémon's shape attribute and its two type attributes. These attributes were selected as they generally affect a Pokémon's silhouette and colors. 

The PixelCNN was trained over 25 epochs with the Adam optimizer with a learning rate of $0.0001$ and a batch size of 32. It uses 7 gated convolutional layers, each with $256$ 3x3 filters. We trained one version of this model for each of our Pixel VQ-VAEs (LowRes, MedRes, HiRes). We repeat the same procedure with a set of VQ-VAEs without our enhancements. All the PixelCNN models have the same architecture and parameters, differing only in the embeddings they use. For complete fairness, we trained a PixelCNN model directly on our training dataset (that is, without using VQ-VAE embeddings) like in \citet{pixel-rnn}. However, the model failed to converge and generated only blank images. We suspect that this is due to the much larger output space in this scenario. For any given pixel there are $256^3$ or over 16 million possibilities. In contrast, even the largest Pixel VQ-VAE has only $K$ (256 in our case) possibilities for any given pixel. 

Table \ref{table:fid_metrics} compares the models in terms of number of parameters and the Fréchet Inception Distance (FID) \cite{fid} (computed on 10,000 generated images). FID is a metric that tells us how close a generated sample is to the training data distribution. This metric was created to compare photo-realistic images and we use it only in the absence of a better metric. Table \ref{table:generated_images} depicts selected outputs from each model. Our first baseline, the VAE, suffers from extreme blurriness with no recognizable images whatsoever, which its FID score reflects. Although the DCGAN achieves an FID score close to our best model (Pixel VQ-VAE HiRes), we see a clear distinction in the generated images. While the GAN does generate interesting shapes with a good spectrum of colors, the generated images no longer retain the pixel art style that we desire. On the other hand, our Pixel VQ-VAE generates interesting shapes and employs color gradients while still retaining the pixel art style. As discussed in the training subsection, the Pixel VQ-VAE HiRes model has the least parameters due to the low value of $L$. While smaller in comparison to the GAN, it must be considered in tandem with the PixelCNN which is responsible for generation. We reiterate that a HiRes VQ-VAE cannot exist without our enhancements. Our Pixel VQ-VAEs have far more parameters than the baselines for the same reason. Although we tried tuning the hyperparameters of the baselines in order to match the number of parameters, we found that this led to very poor convergence. Comparing the Pixel VQ-VAE and the baseline, our Pixel VQ-VAE works better in the LowRes case while in the MedRes case the VQ-VAE beats it out by a small margin. Our Pixel VQ-VAE LowRes and MedRes models are larger than the baselines due to the addition of the PixelSight block and Adapter layers. However, taking into consideration both the quality of the embeddings and the FID scores, our Pixel VQ-VAEs are the best performing models overall. It is also evident that none of the generations appear representative of new Pokémon. Pokémon, due to their high variance, are difficult to generate. Despite our enhancements to the VQ-VAE, a PixelCNN is insufficient on its own to completely solve this task. We provide further examples of random outputs in the appendix.

\subsection{Image Transformations}

\begin{table}[t]
\centering
\begin{tabular}{cccc}
\hline
\textbf{Source} & \textbf{Target} & \textbf{Hand-} & \textbf{Pixel}\\
\textbf{Image} & \textbf{Image} & \textbf{Authored} & \textbf{VQ-VAE}\\
\hline
\includegraphics[]{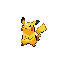} & \includegraphics[]{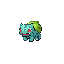} & \includegraphics[]{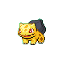} & \includegraphics[]{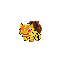}\\
\hline
\includegraphics[]{images/bulbasaur_base} & \includegraphics[]{images/pikachu_base} & \includegraphics[]{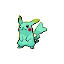} & \includegraphics[]{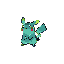}\\
\hline
\end{tabular}
\caption{Sample results for the palette swapping task.}
\label{table:color_swap}
\end{table} 
We consider a palette swapping task due to its popularity in prior work \cite{pokemon2pokemon,serpa-spritesheets-2019,gonzalez-etal}. This involves modifying color palettes of images while retaining their shape and structure. This is used to generate different variations of a particular Pokémon, perhaps with different attributes associated with it. We note that this is a highly subjective task with no real metrics apart from an eye test. Table \ref{table:color_swap} shows an example of a hand-crafted \cite{poke-colors} color swap and one generated by our HiRes Pixel VQ-VAE. To obtain these color swapped images, we first compile frequency statistics for each image's encodings. We then swap the encodings used by the images while retaining the same frequency statistics to obtain color-swapped images. We describe this in more detail in the appendix. We note that despite not performing any special training procedure for this task, our model is still capable of arriving at a reasonable approximation of a human-authored image without ever seeing one. While we leave the implementation to future work, we believe that achieving near-human results on this task would be doable. 

\section{Future Work \& Conclusions}

Our study has shown that the Pixel VQ-VAE meets or exceeds the performance of all baselines. This is a promising start for future approaches into pixel art. We further identify some limitations and extensions to our work. First, tasks like image generation do not have specialized metrics for pixel art. While a full user study is one form of evaluation, developing a metric is another avenue for future work. Additionally, while the VQ-VAE worked quite well, there are several models that build on it that could be used instead \cite{vqvae-2,larsen2016autoencoding}. Finally, there are a number of other downstream tasks the embeddings could be used in, such as image classification or entity representations in reinforcement learning environments. 

In this paper we introduced the Pixel VQ-VAE to represent pixel art as a set of discrete encodings. To the best of our knowledge this work is the first to leverage the encoding-pixel correspondence of VQ-VAEs for such a task. We proved the strength of the Pixel VQ-VAE against several baselines, illustrated the superior embedding space learned by our model and further demonstrated the performance of our model on downstream tasks. Our hope is for this work to act as a starting point for future forays into pixel art.

\section*{Acknowledgments}
We acknowledge the support of the Natural Sciences and Engineering Research Council of Canada (NSERC) and Alberta Machine
Intelligence Institute (Amii).

\bibliography{aaai22}

\appendix

\begin{table*}[t]
\centering
\begin{tabular}{cccccccc}
\hline
\textbf{Original} & \textbf{VAE} & \textbf{Pixel VAE} & \textbf{VQ-VAE} & \textbf{Pixel VQ-VAE} & \textbf{VQ-VAE} & \textbf{Pixel VQ-VAE} & \textbf{Pixel VQ-VAE}\\
\textbf{} & \textbf{} & \textbf{} &\textbf{LowRes} & \textbf{LowRes} & \textbf{MedRes} & \textbf{MedRes} & \textbf{HiRes}\\
\hline\\
\includegraphics[]{images/pokemon_sample_input} & \includegraphics[]{images/pokemon_sample_vae} & \includegraphics[]{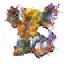} & \includegraphics[]{images/pokemon_sample_vq_vae_small} & \includegraphics[]{images/pokemon_sample_pixel_vq_vae_small} & \includegraphics[]{images/pokemon_sample_vq_vae_medium} & \includegraphics[]{images/pokemon_sample_pixel_vq_vae_medium} & \includegraphics[]{images/pokemon_sample_pixel_vq_vae_large} \\
\includegraphics[]{images/magikarp_sample_input} & \includegraphics[]{images/magikarp_sample_vae} & \includegraphics[]{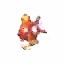} & \includegraphics[]{images/magikarp_sample_vq_vae_small} & \includegraphics[]{images/magikarp_sample_pixel_vq_vae_small} & \includegraphics[]{images/magikarp_sample_vq_vae_medium} & \includegraphics[]{images/magikarp_sample_pixel_vq_vae_medium} & \includegraphics[]{images/magikarp_sample_pixel_vq_vae_large} \\
\hline
\end{tabular}
\caption{Test set reconstruction comparison of the Pixel VAE.}
\label{table:pixel_vae_outputs}
\end{table*} 

\begin{table}[t]
\centering
\begin{tabular}{l|r|r}
\hline
\textbf{Model} & \textbf{MSE} & \textbf{SSIM}\\ 
\hline
VAE & 0.01815 & 0.64272 \\ 
Pixel VAE & 0.01257 & 0.74339  \\ 
VQ-VAE LowRes & 0.01076 & 0.60198 \\ 
Pixel VQ-VAE LowRes & 0.01040 & 0.78669 \\ 
VQ-VAE MedRes & 0.00584 & 0.63973 \\ 
Pixel VQ-VAE MedRes & 0.00421 & \textbf{0.91210} \\ 
Pixel VQ-VAE HiRes & \textbf{0.00070} & 0.82967 \\
\hline
\end{tabular}
\caption{Test set reconstruction metrics. MSE: Lower is better. SSIM: Higher is better. }
\label{table:pixel_vae_metrics}
\end{table} 

\section{Appendix: Pixel VAE}

In addition to our Pixel VQ-VAE, we also tried several enhancements to the baseline VAE, including the addition of a PixelSight block. Like the baseline described in the paper, this model has a mirrored encoder-decoder architecture with 4 convolutional blocks. We add an initial PixelSight block to the start of the encoder and another PixelSight block with transposed convolutions at the end of the decoder. In addition, we heavily modified the training objective. First, we modified the reconstruction loss to use a combination of the Mean Squared Error (MSE) and the Structural Similarity Index (SSIM). We reasoned that MSE, a per-pixel loss function, did not contribute as much to the structure of the reconstructed image. As such including the SSIM which focuses purely on the structure of the image might help in better reconstructions. In an attempt to give a greater focus to the reconstruction loss, we also experimented with a weighted KL-Divergence. After some tuning we found that a weight of 0.1 gave us the biggest improvement in terms of both MSE and SSIM. We trained this model for 25 epochs with a batch size of 64 with the Adam optimizer and a learning rate of 0.0001. Table \ref{table:pixel_vae_metrics} compares the results of these improvements against the other baselines as well as our Pixel VQ-VAE. We see that there is a significant improvement over the baseline VAE, particularly in the SSIM score. However, this performance is still much worse than the VQ-VAEs, especially in terms of the MSE. Table \ref{table:pixel_vae_outputs} gives a visual comparison of the same. 

\section{Appendix: Palette Swap Task}
In this section we go into further detail on how we perform the palette swapping task. For each image, we first compile a map of the encodings present in the image in descending order of the number of their occurrences. We then simply swap the encodings in order. So image A's most frequently used encoding would be image B's most frequently used encoding and vice-versa. We note that this is a simple method of performing this palette swap. There may be better ways of performing this task that take into account edge cases. However, we believe that the examples shown in Table \ref{table:color_swap} display the utility of our representation for this task. We further note that this same methodology can be used to recolor Pokemon to custom colors by simply mapping the encodings of the Pokemon to the encodings corresponding to the desired colors. This does require those colors to be present in our learned encodings, but this aligns well with the limited color palette of pixel art.

\section{Appendix: Generated Outputs}

We include a selection of randomly sampled images generated by the models described in the image generation section of the main paper. All VQ-VAEs use a PixelCNN for generation and all models were trained on the same dataset. Note that the VQ-VAE models learned to mask out the noise while the GAN does not.

\begin{figure*}[t]
    \centering
    \includegraphics[width=0.49\textwidth]{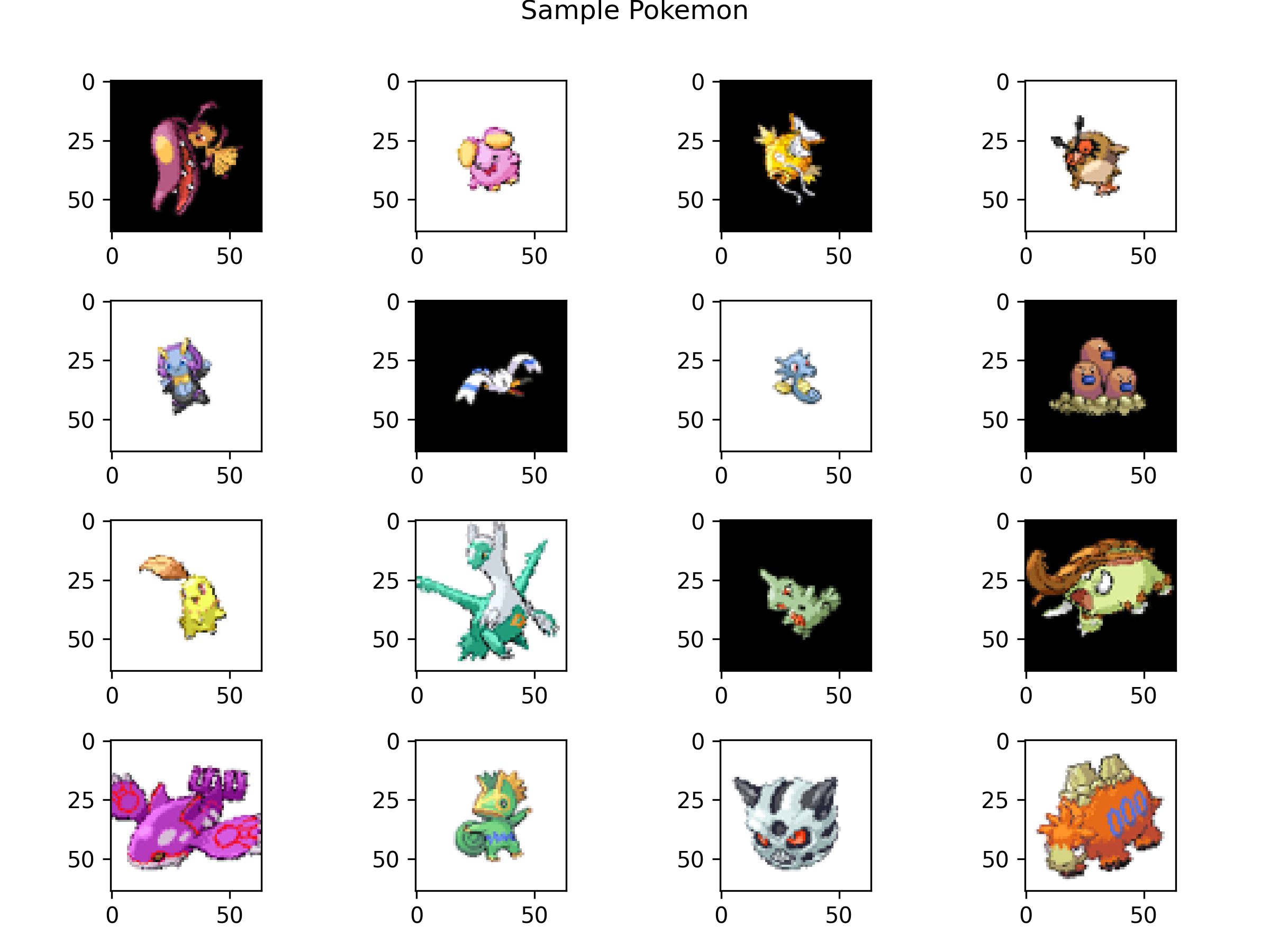}
    \includegraphics[width=0.49\textwidth]{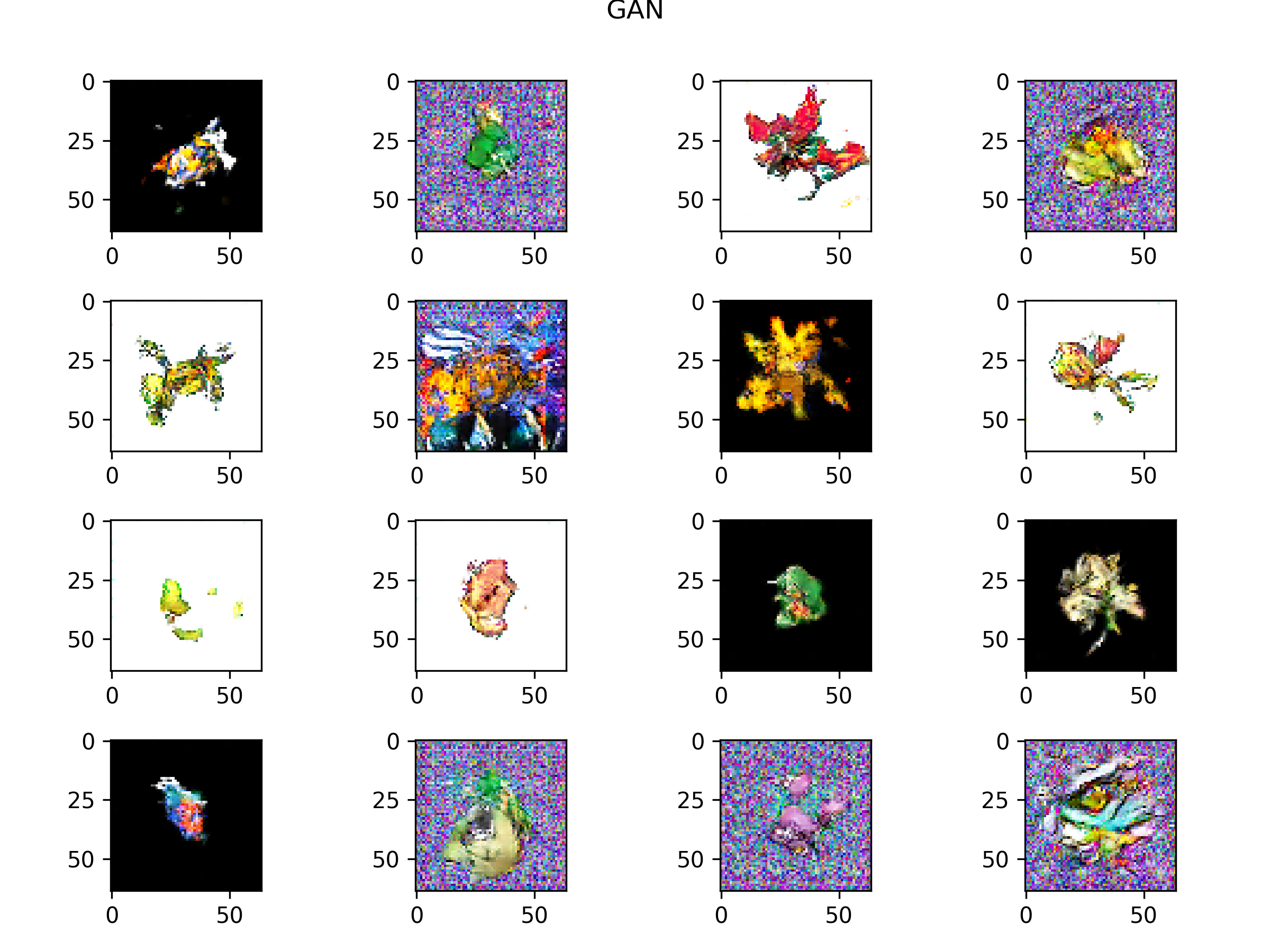}
    \includegraphics[width=0.49\textwidth]{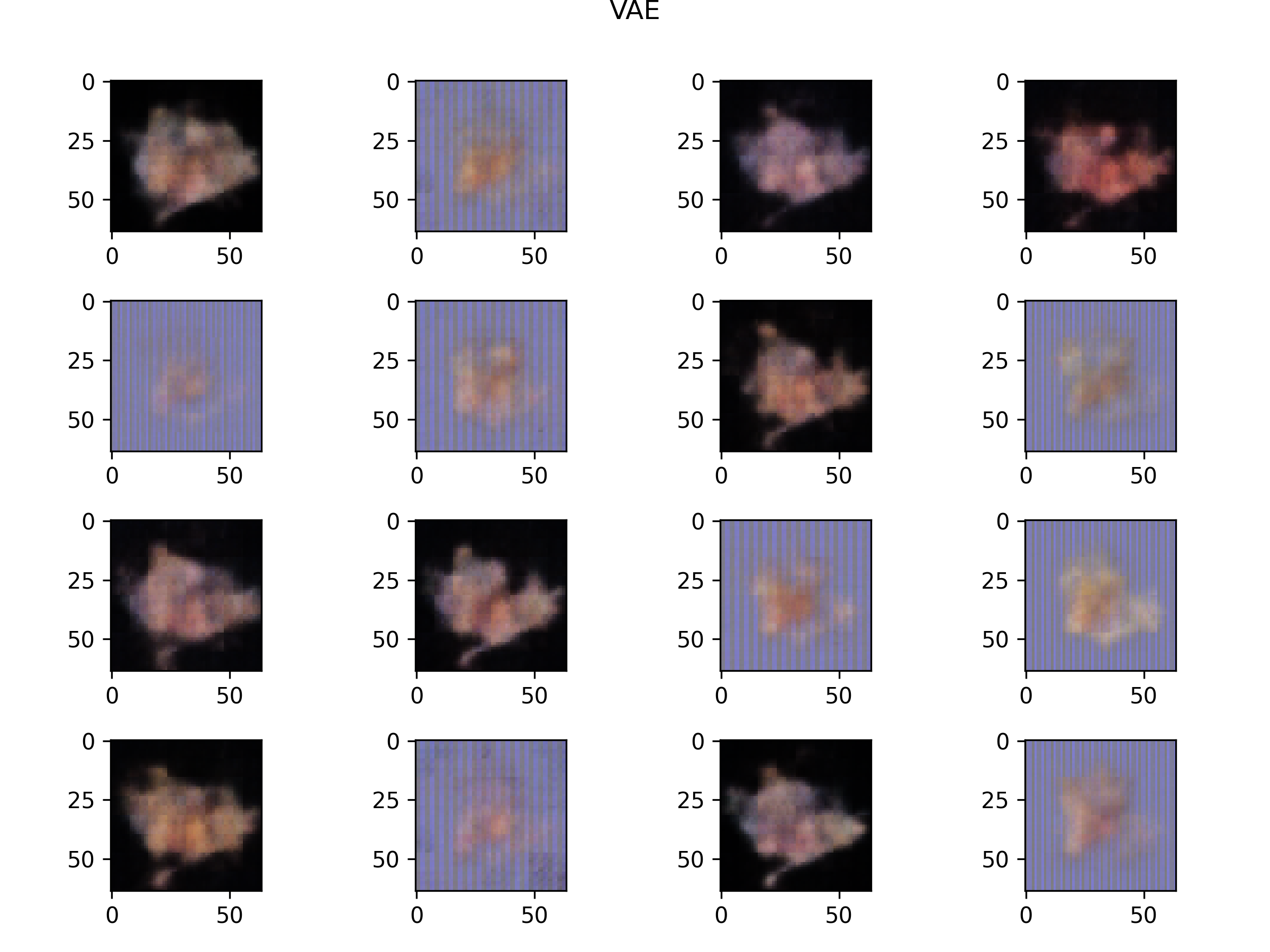}
    \includegraphics[width=0.49\textwidth]{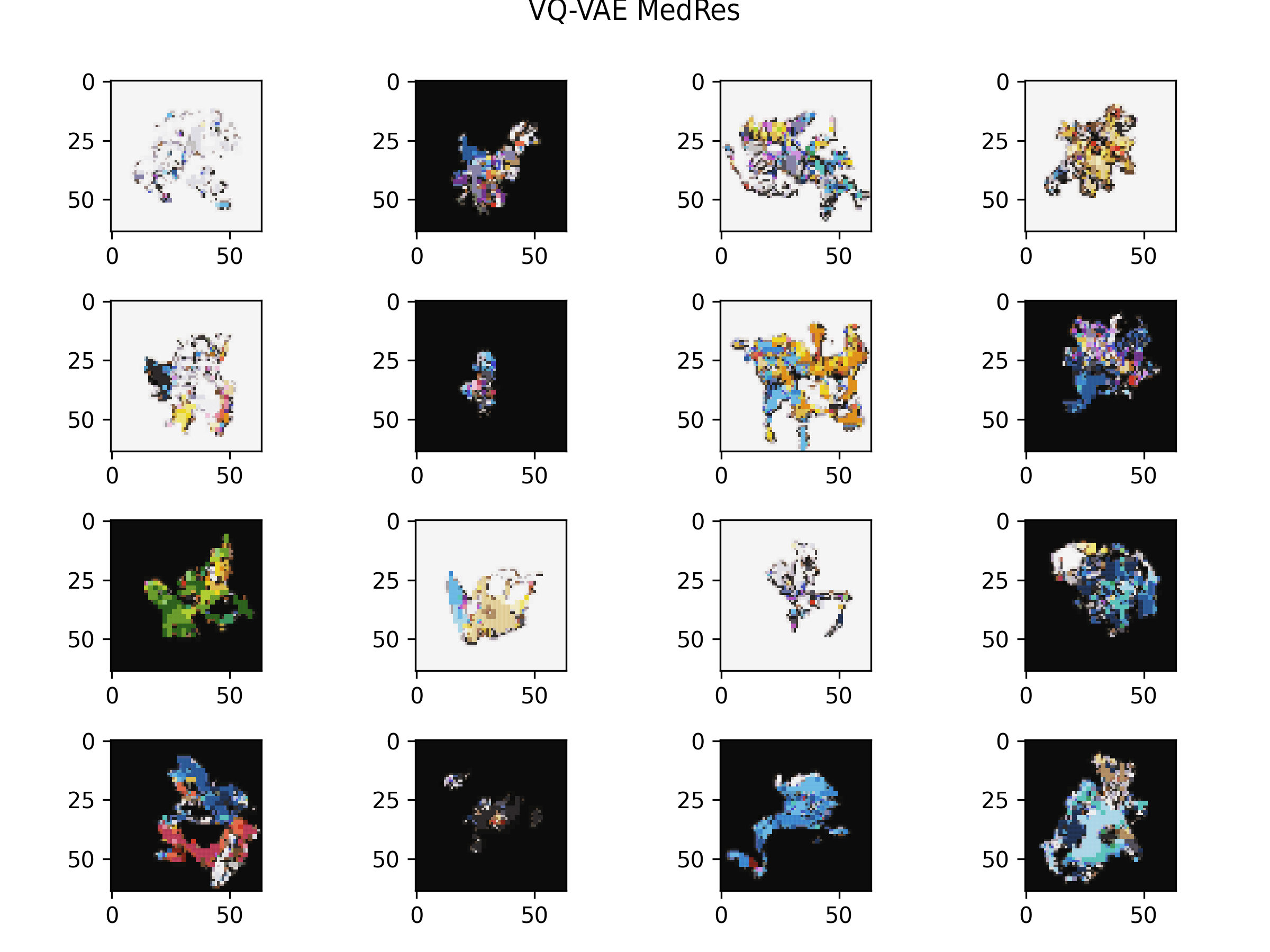}
    \includegraphics[width=0.49\textwidth]{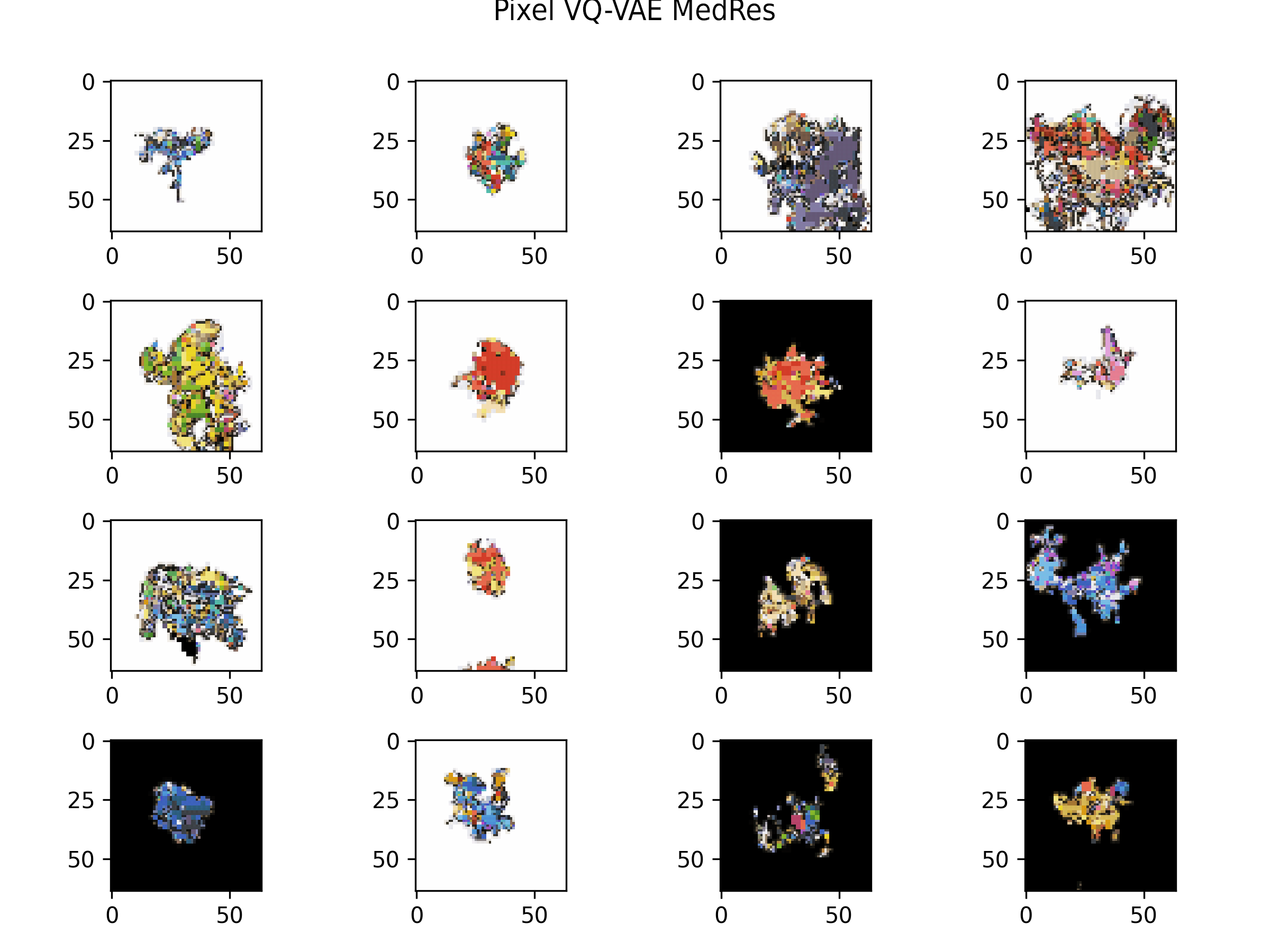}
    \includegraphics[width=0.49\textwidth]{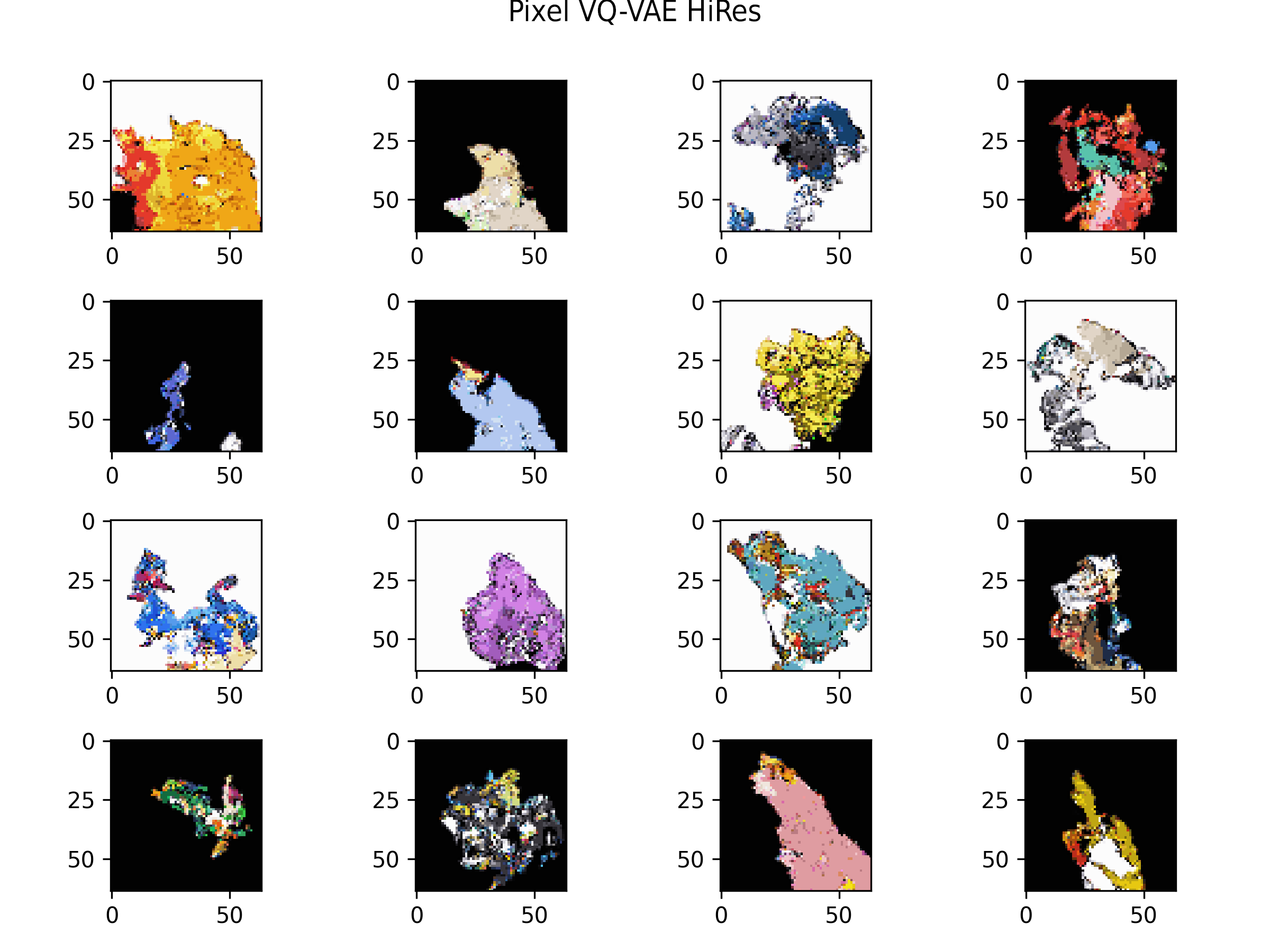}
    \caption{A comparison of the generated images from our Pixel VQ-VAE against several baselines. We do not include the LowRes versions of the models due to space constraints. Samples are randomly pulled from 10,000 generated images.}
    \label{figure:pokemon_generated_samples}
\end{figure*}

\end{document}